\documentclass[letterpaper]{article} 

\ifdefined\aaaianonymous
    \usepackage[submission]{aaai2026}  
\else
    \usepackage{aaai2026}              
\fi
\usepackage{booktabs}      
\usepackage{pifont}        
\usepackage{xcolor}        
\usepackage{graphicx}      
\usepackage{times}  
\usepackage{amssymb}  
\usepackage{pifont}
\usepackage{multirow} 
\usepackage{helvet}  
\usepackage{courier}  
\usepackage[hyphens]{url}  
\usepackage{graphicx} 
\usepackage{natbib}  
\usepackage{caption} 
\usepackage{booktabs}
\usepackage{amsmath}
\usepackage{algorithm}
\usepackage{algorithmic}

\usepackage{newfloat}
\usepackage{listings}
\DeclareCaptionStyle{ruled}{labelfont=normalfont,labelsep=colon,strut=off} 
\floatstyle{ruled}
\newfloat{listing}{tb}{lst}{}
\floatname{listing}{Listing}

\ifdefined\aaaianonymous
    \title{LOC: A General Language-Guided Framework for \\Open-Set 3D Occupancy Prediction}
\else
    \title{LOC: A General Language-Guided Framework for \\Open-Set 3D Occupancy Prediction}
\fi

\author{
   Yuhang Gao$^1$, Xiang Xiang$^{1,2 *}$, Sheng Zhong$^1$, Guoyou Wang$^1$
}
\affiliations{
    \textsuperscript{\rm 1} Nat. Key Lab of Multi-Spectral Info Intelligent Processing Tech, School of AI $\&$ Automation,\\
     \textsuperscript{\rm 2} HUST AI $\&$ Visual Learning Lab, School of Computer Science and Technology,\\

    Huazhong University of Science and Technology (HUST), China
}

\usepackage{bibentry}

\begin{document}

\maketitle

\begin{abstract}
Vision-Language Models (VLMs) have shown significant progress in open-set challenges. However, the limited availability of 3D datasets hinders their effective application in 3D scene understanding. We propose LOC, a general language-guided framework adaptable to various occupancy networks, supporting both supervised and self-supervised learning paradigms.
For self-supervised tasks, we employ a strategy that fuses multi-frame LiDAR points for dynamic/static scenes, using Poisson reconstruction to fill voids, and assigning semantics to voxels via K-Nearest Neighbor (KNN) to obtain comprehensive voxel representations. To mitigate feature over-homogenization caused by direct high-dimensional feature distillation,  we introduce Densely Contrastive Learning (DCL). DCL leverages dense voxel semantic information and predefined textual prompts. This efficiently enhances open-set recognition without dense pixel-level supervision, and our framework can also leverage existing ground truth to further improve performance.
Our model predicts dense voxel features embedded in the CLIP feature space, integrating textual and image pixel information, and classifies based on text and semantic similarity. Experiments on the nuScenes dataset demonstrate the method's superior performance, achieving high-precision predictions for known classes and distinguishing unknown classes without additional training data.

\end{abstract}

\ifdefined\aaaianonymous
\else
\fi

\ifdefined\aaaianonymous

\begin{quote}\begin{scriptsize}\begin{verbatim}

\end{verbatim}\end{scriptsize}\end{quote}
\fi
\section{Introduction}

3D occupancy prediction is crucial for inferring spatial layouts and identifying objects within a 3D voxel grid, enabling safe and independent navigation in complex environments. Previous works in 3D occupancy prediction primarily focused on a closed-set assumption, relying on supervised training with benchmark datasets tailored for specific categories and tasks  \cite{hou2024fastocc,wang2024panoocc,zhang2023occformer,li2023voxformer,huang2021bevdet}. 
\begin{figure}[t]
  \centering
  \includegraphics[width=1\linewidth]{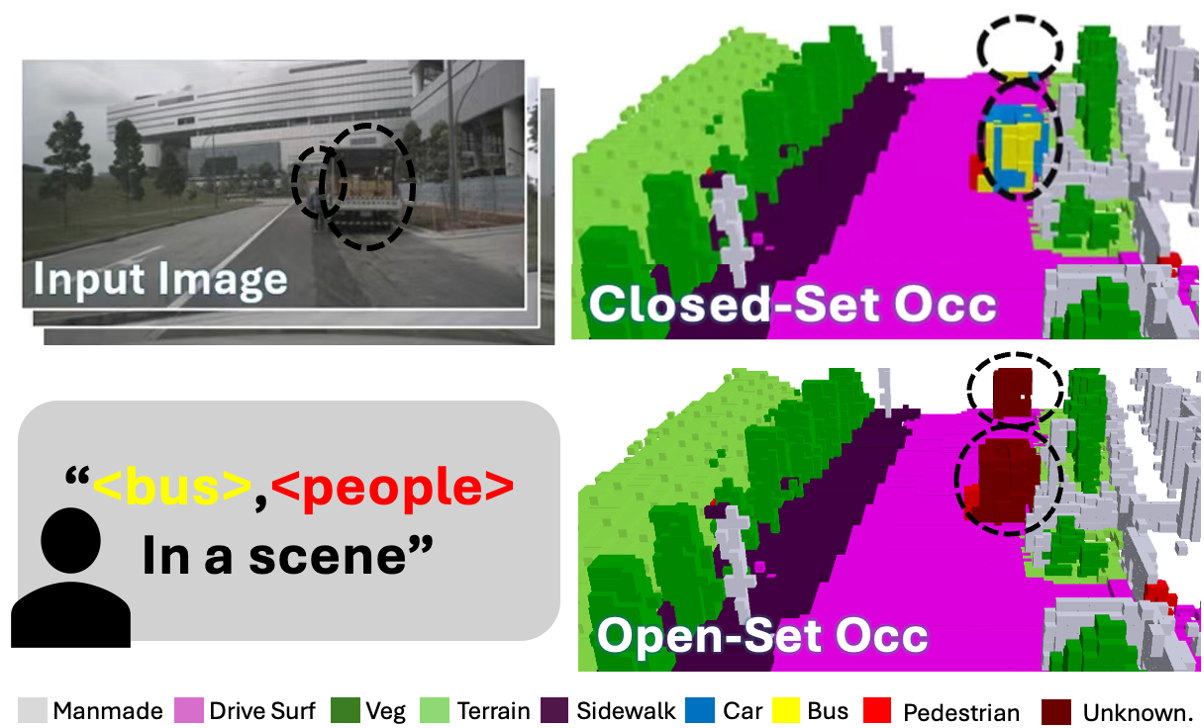} 
  \caption{Given a set of images containing a previously-unknown object (left), the closed-set occupancy model classifies the voxels belonging to that object as either a known category or as free (center, black circle). Our goal is to predict known classes and identify unknown classes.
}
  \label{fig：introduction}
\end{figure}
Consequently, these approaches reveal a significant gap between model capabilities and the demands of real-world autonomous driving scenarios, as solely relying on closed-set training data limits a model's ability to cover all potential object categories an autonomous vehicle might encounter.
\par
While VLMs, trained on vast image-text pairs, offer a promising solution for open-set recognition by enabling 2D-to-3D knowledge transfer, challenges persist. Recent VLM extensions to 3D open-vocabulary tasks \cite{boeder2024langocc,vobecky2023POP3D,zheng2025veon,tan2023ovo} highlight difficulties, particularly in acquiring high-quality dense 3D occupancy representations for self-supervised learning. The process of projecting sparse LiDAR points to 2D images for feature extraction often results in non-dense feature maps. The process of projecting sparse LiDAR points to 2D images for feature extraction often results in non-dense feature maps. Naive multi-frame LiDAR fusion is limited to static scenes and often yields sparse point clouds with voids, leading to erroneous labels. Traditional densification methods, such as nearest neighbor assignment or direct pixel-level feature distillation from CLIP, frequently suffer from feature over-homogenization, where spatially proximate yet semantically distinct voxels are assigned similar features. This can result in missing details, difficulty differentiating adjacent objects, and, when extended to high-dimensional feature distillation, can solidify errors and introduce substantial storage and training overhead. Furthermore, existing VLM frameworks often rely on the assumption that segmentation networks recognize all image categories, prioritizing semantic relationships over novel object discovery, despite many real-world objects being absent from training datasets.

To address these limitations, we propose LOC: a general language-guided framework for open-set 3D occupancy prediction. Our framework is designed for high adaptability across various occupancy networks and robustly supports both supervised and self-supervised learning paradigms.
For self-supervised tasks, we introduce a robust densification strategy. This strategy involves separately fusing multi-frame LiDAR points for dynamic objects and static scenes, then employing Poisson reconstruction to effectively fill voids and obtain comprehensive voxel representations. Subsequently, semantics are assigned to these voxels via KNN interpolation. To circumvent issues like feature over-homogenization and error solidification often caused by direct high-dimensional feature distillation, we introduce Dense Contrastive Learning. DCL leverages dense voxel semantic information and predefined textual prompts, efficiently enhancing open-set recognition without dense pixel-level supervision, and avoiding the storage and training overhead associated with dense high-dimensional features. Furthermore, we emphasize that for 3D occupancy tasks in self-driving, identifying new objects with limited labels is crucial, and thus our framework can also effectively leverage existing ground truth data to further improve performance.

Our model predicts dense voxel features embedded within the CLIP feature space, effectively integrating textual and image pixel information, and performs classification based on text and semantic similarity. Experiments on the nuScenes dataset demonstrate the superior performance of our method, achieving high-precision predictions for known classes and remarkably distinguishing unknown classes without requiring additional training data. We hope this work will inspire further thinking and exploration into open-set 3D occupancy prediction models.

Our contributions are summarized as follows:

\begin{itemize}
    \item We propose LOC, a novel language-guided framework for open-set 3D occupancy prediction, which is the first exploration, according to our knowledge.
    \item We propose Dense Contrastive Learning, a novel method that leverages dense voxel semantic information and textual prompts, efficiently enhancing open-set recognition.
    \item We established comprehensive evaluation baselines on the nuScenes dataset, upon which our framework achieved high-precision predictions for known classes and distinguished unknown classes.
\end{itemize}

\section{Related Work}
\subsection{3D Occupancy Prediction}

Generating dense representations of a 3D scene’s geometry and semantics from visual data has become a pivotal task in computer vision. Traditionally, this understanding has been accomplished using high-precision LiDAR sensors and evaluated on dedicated LiDAR benchmarks. Although LiDAR provides accurate depth information, its inherent sparsity limits comprehensive scene understanding. Recent advances in 3D occupancy prediction leverage multi-camera Bird's-Eye-View (BEV) projections to capture global scene representations. Methods like BevDet aggregate multi-view image features into BEV space \cite{huang2022bevdet4d,huang2022bevpoolv2,li2023fb,li2022bevformer,yang2023bevformer,sodano2024cvpr,huang2021bevdet}, while TPVFormer \cite{huang2023tri} proposes a tri-perspective view. FlashOCC \cite{yu2023flashocc} and SparseOCC \cite{liu2023fully} focus on developing efficient and deployment-friendly occupancy prediction models.
Despite these improvements, such methods typically trained on a closed set of predefined classes, lacking the ability to handle unknown categories. This limitation restricts their applicability in dynamic real-world scenarios where new objects or classes may appear.

\subsection{Open-Set Segmentation} 

Open-set or anomaly segmentation extends the OOD task by attempting to predict whether each pixel in an image belongs to an unknown class. This approach aims to identify not only pixels from known classes but also detect those that fall outside the distribution of the training data. Such tasks are particularly important in autonomous driving, where effectively detecting objects of novel categories is critical. One straightforward approach is to apply a threshold to the softmax outputs, as seen in MSP \cite{hendrycks2016baseline}. However, for unknown samples, closed-world models often exhibit overconfidence in their predictions. Recent research has also extended these to 3D LiDAR point clouds. Vision-Language Models like CLIP are increasingly used for anomaly segmentation, with methods exploring prompt learning and training-free OOD detection \cite{zhong2022regionclip,rao2022denseclip,oseg}.

\subsection{Open-Vocabulary Segmentation} 
The task has also been extended into the 3D domain. For example, OpenScene \cite{Peng2023OpenScene} achieves this by aligning CLIP-derived features with point cloud features, providing a foundation for 3D open-vocabulary segmentation. POP-3D \cite{vobecky2023POP3D} further uses LiDAR supervision to develop an open-vocabulary model; however, it suffers from severe sparsity issues. VEON \cite{zheng2025veon} introduces a vocabulary-enhanced occupancy framework trained with LiDAR supervision, leveraging CLIP features for open-vocabulary prediction and addressing depth ambiguities via enhanced depth mode.

Most existing works operate under the assumption that class-agnostic segmentation networks can effectively detect all objects within a scene, focusing heavily on improving classification accuracy over segmentation quality. This emphasis on classification often results in an over-reliance on a large set of categories during training, leading models to learn semantic relationships between known classes rather than enhancing their capacity to identify unknown objects. They neglect the fact that training sets cannot cover all possible object categories in practice. In autonomous driving, the ability to detect unknown classes is more critical than achieving detailed semantic classifications.

 \begin{figure*}[ht]
  \centering
  \includegraphics[width=1\linewidth]{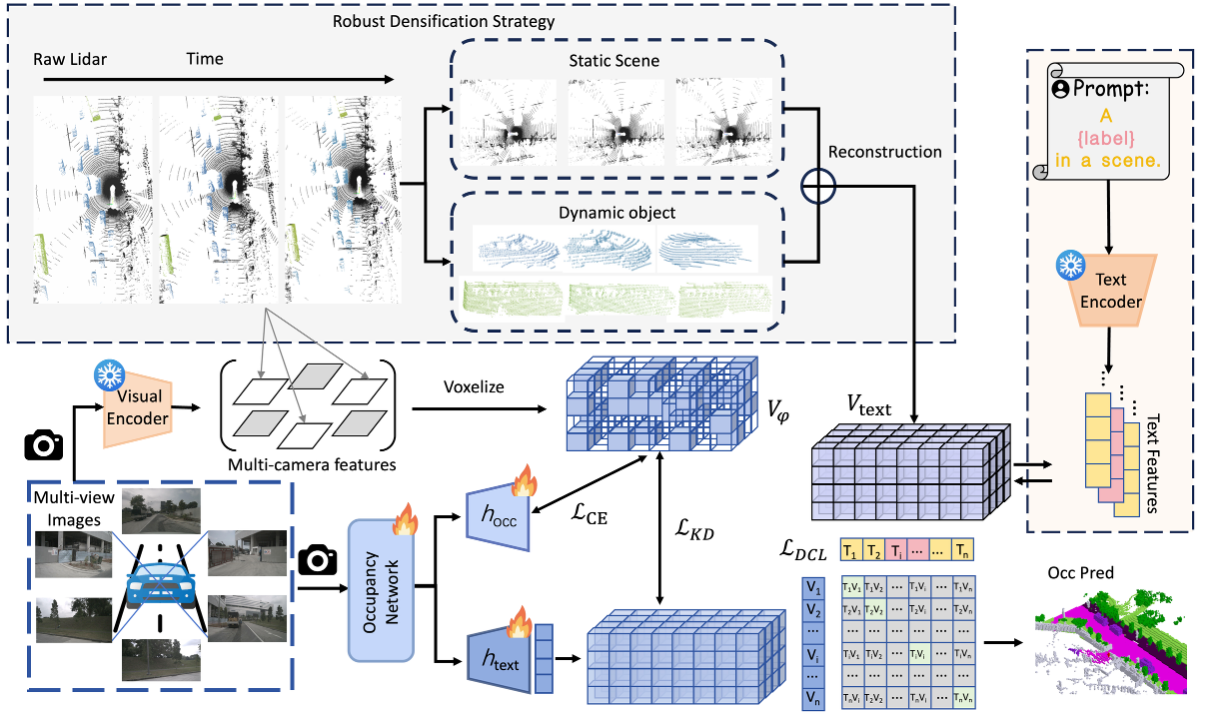} 
 \caption{\textbf{Architecture of the proposed model.} Input images are first transformed into 3D voxel features by the occupancy network M. Features are then fed into two parallel decoding heads: Occupancy Head performs occupancy prediction, where the dense occupancy information obtained from the Robust Densification Strategy serves as the primary supervision signal for the Occupancy Head; Language Head aligns voxel features with CLIP text embeddings using distillation loss and DCL loss. }

  \label{fig：model}
\end{figure*}
 \section{Methodology}
 This study proposes the LOC general framework, aiming to address the challenge of open-set 3D occupancy prediction in autonomous driving. The framework first achieves 2D-to-3D knowledge transfer by efficiently distilling rich semantic information from pre-trained 2D vision-language models into 3D voxel space (an overview of our approach is illustrated in Fig \ref{fig：model}). Subsequently, to overcome issues of sparse 3D data and voids, we designed a Robust Densification Strategy, which generates high-quality dense 3D occupancy representations through dynamic-static scene separation and Poisson reconstruction. Crucially, we introduce the DCL component. While avoiding dense pixel-level supervision, DCL effectively enhances open-set recognition capabilities and addresses feature over-homogenization issues through textual prompts and contrastive learning. Furthermore, DCL possesses the potential to leverage existing 3D occupancy ground truth to further reinforce its performance. Finally, the LOC model combines the outputs of the occupancy head and language head to achieve precise recognition and classification of both known and unknown categories.

 \subsection{2D to 3D Mapping}
 Given RGB images, we use OpenSeg \cite{oseg} to extract pixel-level features. For precise 2D-to-3D mapping, 3D LiDAR points are projected onto multi-camera image planes, with features extracted via bilinear interpolation for each projected point.
 Features from multiple camera views for each 3D point are aggregated using a pooling function, then assigned to corresponding voxels. If multiple points fall into the same voxel, their features are combined via pooling, generating a sparse voxel feature tensor. This process accurately projects 2D image features onto the 3D voxel feature map, generating a sparse voxel feature tensor \( V_{\psi} \in \mathbb{R}^{H \times W \times D \times C_o} \), where \( H, W, D \) denote the dimensions of the voxel grid.

\par
Then, we extract semantic features for each voxel from the 3D voxel feature map $V = \mathcal{M}(I) \in \mathbb{R}^{H \times W \times D \times C_v}$ output by the occupancy network $\mathcal{M}$. This feature is passed to two different prediction heads. 
\paragraph{Occupancy Head \(h_{\text{occ}}\). }
The occupancy head serves as a classifier that evaluates the occupancy status of each voxel. In training, we observed that using a multi-class classification task to train the occupancy head, compared to a pure binary classification task (only distinguishing 'occupied' and 'free'), improved its performance in open-set scenarios. During training, voxel features \( V \) are processed through \(h_{\text{occ}}\) to produce a multi-class prediction tensor \( P^{\text{3D}}  \), where each voxel location is associated with a probability distribution over \( K \) classes, supervised by a cross-entropy loss \( L_{\text{CE}} \). The output tensor can be expressed as:
\begin{equation}
P^{\text{occ}} =  h_{\text{occ}}(V) \in \mathbb{R}^{H \times W \times D \times K}
\label{3D}
\end{equation}

\par
\paragraph{Language Head \(h_{\text{text}}\).}  

The Language Head is designed to refine the semantic understanding of occupied voxels. It takes as input the features from voxels identified as occupied by the Occupancy Head. This head, composed of multi-layer perceptrons, maps these voxel features into a semantic feature space aligned with text embeddings:
    
\begin{equation}
V_{\text{text}} = h_{\text{text}}(V) \in \mathbb{R}^{H \times W \times D \times C_o}
\end{equation}

To ensure that the dense language features $V_{\text{text}}$ maintain consistency with the initial sparse semantic information derived from 2D projections, we introduce a knowledge distillation (KD) loss. This loss encourages $V_{\text{text}}$ to align with the sparse voxel feature tensor $V_{\psi}$ at locations where $V_{\psi}$ is occupied. Specifically, for voxels $s$ where $V_{\psi}(s)$ is occupied, we minimize the cosine distance between the corresponding features from $V_{\text{text}}$ and $V_{\psi}$:

\begin{equation}
\mathcal{L}_{\text{KD}} = \frac{1}{N_{\text{s}}} \sum_{s \in V_{\psi}(s)} \left( 1 - \text{sim}(V_{\text{text}}(s), V_{\psi}(s)) \right)
\end{equation}

\noindent where $N_{\text{s}}$ is the number of occupied voxels in $V_{\psi}$, and $\text{sim}(\cdot,\cdot)$ denotes cosine similarity.

\subsection{Robust Densification Strategy}
We observed that networks only supervised by sparse voxel features $V_{\psi}$ face significant challenges in generating dense occupancy representations. Following the core principles of existing methods\cite{tian2024occ3d,kazhdan2006poisson}, we designed a Robust Densification Strategy to produce high-quality dense occupancy voxels.

Since the point cloud distribution of dynamic objects with non-negligible velocities exhibits spatiotemporal variations in the world coordinate system, it is imperative to individually extract and reconstruct these dynamic entities when performing per-frame analysis in 3D space.
\paragraph{Dynamic-Static Separation.}
Specifically, for each frame $t$, raw LiDAR point clouds \(P_t^{\text{LiDAR}}\) are segmented into movable object points (identified via 3D bounding boxes) and static scene points, with ego-vehicle points filtered out. Dynamic objects across frames are aggregated using consistent tracking IDs, while static segments form a global point cloud $P_t^{\text{LiDAR}}$. We utilize consistent tracking IDs for the same object across the time series to identify and aggregate dynamic objects across frames. All aggregations are transformed to the first frame's LiDAR coordinate system.

The complete scene point cloud for frame $t$ is obtained by re-projecting aggregated clouds to the current frame's LiDAR coordinate system:
\[
P_t=\left[T_{s\to t}(P_s),T_{o\to t}(P_o)\right]
\]

\noindent where \(T_{s \to t}\) and \(T_{o \to t}\) transform static and dynamic point clouds, respectively, yielding the dense \(P_t\) for voxelization.
This final $P_t$ is the dense point cloud used for voxelization.
\paragraph{Poisson Reconstruction}.
Multi-frame fused \(P_t\) can be reconstructed into a triangular mesh via Poisson Surface Reconstruction \cite{kazhdan2006poisson} to enhance spatial continuity and fill residual voids. Voxelizing this mesh generates a dense 3D occupancy grid \(V^D \in \mathbb{R}^{H \times W \times D}\).Since \(V^D\) only contains occupancy information without semantic categories, we employ K-Nearest Neighbors from the original point cloud, resulting in \(V^{\hat{D}}\).

\subsection{Dense Contrastive Learning }

A direct approach, which involves assigning features to occupied voxels in the dense voxel grid $V^D$ from sparse voxel features $V_{\psi}$ via a Nearest Neighbor algorithm and then directly performing dense distillation, introduces several challenges. This method often leads to feature over-homogenization, where spatially proximate but semantically distinct voxels are assigned similar features, resulting in a loss of critical detail and a reduced ability to differentiate adjacent objects (see Ablation Study). Furthermore, such direct distillation can solidify errors, propagating and amplifying inaccuracies from the source features into the target representation. The storage and processing of dense high-dimensional features also incur significant storage and training overhead.
To overcome those limitations, we propose an innovative Dense Contrastive Learning method. The core idea of DCL is to enhance the model's open-set recognition capabilities efficiently, without requiring dense pixel-level supervision, by establishing a contrastive relationship between voxel features and their corresponding textual prompts. This method utilizes a dual-head architecture, where the Language Head is responsible for generating semantic features for voxels and is integrated with the DCL mechanism to boost open-set capabilities.

\paragraph{Text Prompt Construction}
For contrastive learning, we construct a set of predefined textual prompts. We map the original nuScenes categories to more fine-grained ones(see Appendix). This includes prompts for known categories (e.g., "a car in a scene", "a person in a scene", 'other' and ‘free' excluded). We use the CLIP text encoder to extract the text embedding for the class prompt, denoted as  \( E = \{ e_1, e_2, \dots, e_{k} \} \).

DCL operates by optimizing a contrastive loss function designed to maximize the similarity between a voxel's language feature $f_v$ and its corresponding correct text embedding $e_{\text{pos}(v)}$, while minimizing similarity with irrelevant text embeddings. We adopt a variant of the InfoNCE loss:

\begin{equation}
\mathcal{L}_{\text{DCL}} = -\frac{1}{N_v} \sum_{v \in V^{\hat{D}}} \log \frac{\exp\left(\text{sim}(f_v, e_{\text{pos}(v)}) / \tau_1\right)}{\sum_{k \in K} \exp\left(\text{sim}(f_v, e_k) / \tau_1\right)}
\end{equation}

\noindent where $N_v$ is the number of occupied voxels in $V^{\hat{D}}$, $v$ represents a voxel position in the voxel grid, and $f_v \in \mathbb{R}^{C_{\text{o}}}$ is the language feature for that voxel. $e_{\text{pos}(v)}$ is the positive sample text embedding corresponding to voxel $v$, $\text{sim}(\cdot, \cdot)$ denotes cosine similarity, and $\tau_1$ is the temperature parameter. Through this mechanism, the model learns to align voxel features with correct semantic concepts.

Directly using cosine similarity for text feature supervision, however, can lead to significant performance degradation due to class imbalance issues. In contrast, our Densely Contrastive Learning (DCL) approach is designed to robustly handle such complexities and improve feature discrimination.
Ablation experiments further validate the effectiveness of DCL in improving model performance and open-set recognition capabilities.
\paragraph{Reinforcing DCL with 3D Occupancy GT.}
Despite DCL's demonstrated ability to enhance open-set recognition through contrastive learning in the absence of dense pixel-level supervision, we recognize that current 3D occupancy ground truth (GT) generation methods are progressively maturing, and the 3D occupancy prediction task inherently aims to maximize the utilization of limited annotated data. Therefore, our DCL component can further leverage this available 3D occupancy GT information to reinforce its performance. Specifically, by using the occupancy GT as $V^{\hat{D}}$, DCL can acquire more precise semantic supervision, thereby not only improving the recognition accuracy of known classes but also indirectly optimizing its capability to distinguish unknown entities under open-set conditions.

\subsection{Open-set Prediction.}
The final loss \( \mathcal{L} \) is a weighted sum of the distillation loss \( \mathcal{L}_{\text{KD}} \) , DCL loss  \( \mathcal{L}_{\text{DCL}} \) and the cross-entropy loss \( \mathcal{L}_{\text{CE}} \):

\begin{equation}
\mathcal{L} = \mathcal{L}_{\text{CE}} + \lambda_1 \mathcal{L}_{\text{KD}} + \lambda_2 \mathcal{L}_{\text{DCL}}
\end{equation}

\noindent where \( \lambda \) is the weighting factor that controls the relative contributions of the distillation and cross-entropy losses.During inference, the occupancy head determines whether a voxel is occupied. If a voxel is occupied, the language head is then used to perform semantic prediction.

To obtain classification probabilities for each voxel, we compute the cosine similarity between each voxel’s  feature \( f _v\) and the text embedding \( E \). The similarities are then converted into classification probabilities using the Softmax function, with a scaling factor \( \tau_2 \) applied as a divisor. The resulting probability  for class \( k
\) is given by:

\begin{equation}
P_k^\text{text} = \frac{\exp \left( \text{sim}(f_v, e_k) / \tau_2 \right)}{\sum\limits_{k=1}^{K} \exp \left( \text{sim}(f_v, e_k) / \tau_2\right)}
\end{equation}
\paragraph{Post-Processing for unknown classes.} In practice, we observed that the occupancy head can accurately capture the occupancy status of each voxel. On the other hand, the language head outputs features aligned with text embeddings, providing strong generalization and zero-shot capabilities. However, it is prone to show low confidence for both unoccupied and unknown categories due to the absence of corresponding precise textual features. Therefore, our final prediction combines the outputs of both heads. In fact, for each voxel, we compute the maximum value of the logits from the two heads. The formulas for this process are as follows:

\begin{equation}
S_{\text{occ}} = \max_{k} \left( P_k^{\text{occ}} \right)
\label{eq:occ}
\end{equation}

\begin{equation}
S_{\text{text}} = \max_{k} \left( P_k^{\text{text}} \right)
\label{eq:text}
\end{equation}
 
 Finally, we sum these two maximum values to obtain a unknown class score for that voxel as

\begin{equation}
S_{\text{kn}} = \frac{1}{2} \left( s_{\text{occ}} + s_{\text{text}} \right)
\label{eq:kn}
\end{equation}

If the score is below a predefined threshold \(\delta\) , the voxel is considered
belonging to an unknown class.

\section{Experiments}
\paragraph{Experimental Setup.} We conducted experiments on the nuScenes dataset \cite{caesar2020nuscenes}, treating the `others' category (and additional classes explored later) as unknowns $K_{\text{n}}$ and excluding them during training. Based on the benchmark's 16 semantic classes, we defined predefined prompts mapped to 43 categories (see Appendix), ignoring unknown prompts during inference. We evaluated both closed-set and open-set 3D occupancy segmentation using the Occ3D-nuScenes benchmark \cite{vobecky2023POP3D}. Closed-set performance uses mIoU; open-set evaluation uses AUROC and FPR95 (higher AUROC and lower FPR95 are better).
\begin{table}[h]
  \centering
  \caption{Performance comparison of different approaches in the open-set setting. \( K_{\text{n}} \) represents the set of unknown classes, which are ignored during training but evaluated on the Test Set. This table also presents the performance of different occupancy networks, with a special note that LOC-L is a self-supervised method.}
  \label{tab:main}
  \begin{tabular}{@{}c|c| c c c c c@{}}
    \toprule
    Approach  &Occ Network& mIoU & AUPR\(\uparrow\) & FPR95\(\downarrow\) \\
    
     \midrule
     \multicolumn{5}{c}{\( K_{\text{n}} = \{ \text{others, cons.veh.} \} \)} \\
    Closed-set   &BEVDet& 34.75 & --  & --\\
    MSP          &BEVDet& 34.75 & 72.48 & 81.31 \\
    LogitNorm     &BEVDet &34.45  & 74.25  & 72.42 \\
    MCM \     &BEVDet& 34.75 & 76.26  & 71.06 \\
    {LOC-L(ours)}    &BEVDet& \textbf{18.79} & \textbf{75.35}  & \textbf{70.28} \\
    {LOC-T(ours)}    &TPVFormer& \textbf{29.67} & \textbf{78.30}  & \textbf{64.41} \\ 
    {LOC-B(ours)}    &BEVDet& \textbf{34.99} & \textbf{80.25}  & \textbf{63.99} \\
    {LOC-F(ours)}    &FlashOcc& \textbf{35.10} & \textbf{80.42}  & \textbf{63.83} \\
    \midrule 
     \multicolumn{5}{c}{\( K_{\text{n}} = \{ \text{others, tfc.cone, trailer} \} \)} \\
      Closed-set   &BEVDet& 34.41 & --  & --\\
    MSP           &BEVDet& 34.41 & 74.70 & 80.31 \\
     LogitNorm      &BEVDet&33.17    & 77.06 &71.46\\
     MCM      &BEVDet& 34.41 & 75.77  & 71.41 \\
     {LOC-L}    &BEVDet& \textbf{19.15} & \textbf{76.30}  & \textbf{72.21} \\
    {LOC-T}    &TPVFormer& \textbf{28.81} & \textbf{79.04}  & \textbf{65.73} \\
    {LOC-B}   &BEVDet& \textbf{33.12} & \textbf{81.04}  & \textbf{64.53} \\
    {LOC-F}    &FlashOcc& \textbf{33.36} & \textbf{80.83}  & \textbf{62.38}  \\
   
    \midrule 
     \multicolumn{5}{c}{\( K_{\text{n}} = \{ \text{others, barrier, bus, truck} \} \)} \\ 
      Closed-set   &BEVDet& 33.31 & --  & --\\
    MSP          &BEVDet& 33.31 & 68.81 & 85.38 \\
    LogitNorm      &BEVDet&29.92  &67.49   &80.76  \\
    MCM       &BEVDet& 33.31 & 71.63  & 79.39 \\
    {LOC-L}    &BEVDet& \textbf{19.56} & \textbf{75.11}  & \textbf{78.20} \\
    {LOC-T}    &TPVFormer& \textbf{28.92} & \textbf{75.23}  & \textbf{73.21} \\
    {LOC-B}    &BEVDet& \textbf{33.27} & \textbf{77.92}  & \textbf{71.02} \\
    {LOC-F}    &FlashOcc& \textbf{33.65} & \textbf{78.18}  & \textbf{70.61} \\
       \bottomrule
  \end{tabular}
\end{table}
\par 
\begin{table*}[t]
    \centering
    \caption{3D occupancy prediction performance on the Occ3D-nuScenes occupancy benchmark. We report the mIoU for semantics across different categories, along with per-class semantic IoUs. "Occ GT" indicates whether occupancy ground-truth supervision is required. The lower half of the table specifically presents results for language-driven methods.}
    \label{tab:state}
    \footnotesize
    \begin{tabular}{p{1.9cm}|c |p{0.4cm} p{0.4cm} p{0.4cm} p{0.4cm} p{0.4cm} p{0.4cm} p{0.4cm} p{0.4cm} p{0.4cm} p{0.4cm} p{0.4cm} p{0.4cm} p{0.4cm} p{0.4cm} p{0.4cm} p{0.4cm} p{0.4cm} p{0.6cm}}
    \toprule
         Method & 
         
         \rotatebox{90}{Occ GT} &  
         \rotatebox{90}{others} & 
        \rotatebox{90}{\raisebox{-0.2em}{\textcolor{orange}{\rule{0.5em}{0.5em}}}\, barrier} & 
        \rotatebox{90}{\raisebox{-0.2em}{\textcolor{pink}{\rule{0.5em}{0.5em}}}\, bicycle} & 
        \rotatebox{90}{\raisebox{-0.2em}{\textcolor{yellow}{\rule{0.5em}{0.5em}}}\, bus} & 
        \rotatebox{90}{\raisebox{-0.2em}{\textcolor{blue}{\rule{0.5em}{0.5em}}}\, car} & 
        \rotatebox{90}{\raisebox{-0.2em}{\textcolor{cyan}{\rule{0.5em}{0.5em}}}\, cons. veh.} & 
        \rotatebox{90}{\raisebox{-0.2em}{\textcolor{orange}{\rule{0.5em}{0.5em}}}\, motorcycle} & 
        \rotatebox{90}{\raisebox{-0.2em}{\textcolor{red}{\rule{0.5em}{0.5em}}}\, pedestrian} & 
        \rotatebox{90}{\raisebox{-0.2em}{\textcolor{yellow}{\rule{0.5em}{0.5em}}}\, traffic cone} & 
        \rotatebox{90}{\raisebox{-0.2em}{\textcolor{brown}{\rule{0.5em}{0.5em}}}\, trailer} & 
        \rotatebox{90}{\raisebox{-0.2em}{\textcolor{purple}{\rule{0.5em}{0.5em}}}\, truck} & 
        \rotatebox{90}{\raisebox{-0.2em}{\textcolor{magenta}{\rule{0.5em}{0.5em}}}\, drive. surf.} & 
        \rotatebox{90}{\raisebox{-0.2em}{\textcolor{black}{\rule{0.5em}{0.5em}}}\, other flat} & 
        \rotatebox{90}{\raisebox{-0.2em}{\textcolor{purple}{\rule{0.5em}{0.5em}}}\, sidewalk} & 
        \rotatebox{90}{\raisebox{-0.2em}{\textcolor{green}{\rule{0.5em}{0.5em}}}\, terrain} & 
        \rotatebox{90}{\raisebox{-0.2em}{\textcolor{gray}{\rule{0.5em}{0.5em}}}\, manmade} & 
        \rotatebox{90}{\raisebox{-0.2em}{\textcolor{green}{\rule{0.5em}{0.5em}}}\, vegetation} & 
        \multicolumn{1}{c}{mIoU} \\
    \midrule

    TPVFormer    &\ding{51}& 7.2 & 38.9 & 13.7 & 40.8 & 45.9 & 17.2 & 20.0 & 18.9 & 14.3 & 26.7 & 34.2 & 55.7 & 35.5 & 37.6 & 30.7 & 19.4 & 16.8 & 27.83 \\
    OccFormer    &\ding{51}& 5.9 & 30.3 & 12.3 & 34.4 & 39.2 & 14.4 & 16.5 & 17.2 & 9.3 & 13.9 & 26.4 & 51.0 & 31.0 & 34.7 & 22.7 & 6.8 & 7.0 & 21.93 \\
    
    BEVFormer    &\ding{51}& 9.6 & 47.8 & 24.2 & 48.7 & 54.0 & 20.9 & 28.8 & 27.5 & 26.7 & 32.8 & 38.8 & 81.7 & 40.3 & 50.5 & 52.9 & 43.8 & 37.5 & 39.19 \\
    BEVDet    &\ding{51}& 6.7 & 37.0 & 8.3 & 38.7 & 44.5 & 15.2 & 13.7 & 16.4 & 15.3 & 27.1 & 31.0 & 78.7 & 36.5 & 48.3 & 51.7 & 36.8 & 32.1 & 31.64 \\
    SelfOcc   &\ding{55} &0.0 &0.0 &0.0 &0.0 &9.8 &0.0 &0.0 &0.0 &0.0 &0.0 &7.0 &47.0 &0.0 &18.8 &16.6 &11.9 &3.8 &6.76 \\

    \midrule
                       Veon   &\ding{55}& 0.9 & 10.4 & 6.2 & 17.7 & 12.7 & 8.5 & 7.6 & 6.5 & 5.5 & 8.2 & 11.8 & 54.5 & 0.4 & 25.5 & 30.2 & 25.4 & 25.4 & 15.14 \\
                LangOcc   &\ding{55}& 0.0 & 3.1 & 9.0 & 6.3 & 14.2 & 0.4 & 10.8 & 6.2 & 9.0 & 3.8 & 10.7 & 43.7 & 2.2 & 9.5 & 26.4 & 19.6 & 26.4 & 11.84 \\
                LOC-L  &\ding{55}&-- & 11.2 & 7.8 & 8.5 & 17.2 & -- & 10.8 & 8.5 & 10.1 & 7.9& 12.3 & 55.1 & 8.2 & 30.5 & 35.2 & 30.2 & 28.4 & 18.79 \\
        LOC-T  &\ding{51}&-- & 37.6 & 14.2 & 42.1 & 44.8 & - & 20.5 & 17.9 & 15.7 & 25.3 & 33.2 & 56.8 & 34.7 & 36.3 & 31.8 & 18.9 & 15.3 &29.67\\
               LOC-B  &\ding{51}&-- & 39.4 & 12.7 & 36.8 & 44.0 & -- & 15.4 & 16.6 & 16.8 & 29.5 & 31.1 & 78.2 & 37.3 & 47.4 & 51.2 & 36.8 & 31.6 & 34.99 \\ 
                 LOC-F  &\ding{51}&-- & 39.3 & 12.8 & 37.0 & 44.9 & -- & 15.3 & 17.4 & 18.9 & 27.9 & 31.6 & 78.9 & 36.7 & 48.0 & 51.2 & 35.9 & 30.9 & 35.10 \\
       
    \bottomrule
    \end{tabular}
\end{table*}

\par
\paragraph{Training details and parameters.}
For tasks using BEVDet and FlashOcc as the occupancy network, we employ a ResNet-50 \cite{he2016deep} backbone, with an image resolution of \( 256 \times 704 \) and a 2D feature dimension \( C_o \) set to 768 . For TPVFormer, we follow the settings of POP3D\cite{vobecky2023POP3D}. Training is conducted using the AdamW optimizer \cite{loshchilov2017decoupled} with a learning rate of \( 1 \times 10^{-4} \) and gradient clipping, over a total of 24 epochs. We set \( \lambda_1 = 1 \), \( \lambda_2 = 1 \), temperature \( \tau_1 = 0.5 \), and \( \tau_2 = 0.5 \),for a detailed hyperparameter sensitivity analysis, please refer to Appendix. Most experiments are conducted on 6 NVIDIA GeForce RTX 4090 GPUs.

\par

\label{ablation}
 \begin{table}[h]
  \centering
  \caption{Ablation study on different model components. "Dense" refers to Robust Densification Strategy}
  \label{tab:ablation}
  \setlength{\tabcolsep}{4pt} 
  \small 
  \begin{tabular}{c|c c c c c c c}
    \toprule
    Model & \(h_{\text{occ}}\) & \(\mathcal{L}_{\text{KD}}\) & DCL & Dense & mIoU & AUPR\(\uparrow\) & FPR95\(\downarrow\) \\
    \midrule
    \multirow{4}{*}{LOC-F}
      & \ding{55} & \ding{55} & \ding{51} & - & 12.52 & 30.89 & 95.84 \\
      & \ding{51} & \ding{51} & \ding{55} & - & 8.99 & 71.13 & 88.24 \\
      & \ding{51} & \ding{55} & \ding{51} & - & 25.77 & 74.33 & 85.40 \\
      & \ding{51} & \ding{51} & \ding{51} & - & 35.10 & 80.42 & 63.83 \\
    \midrule
    \multirow{3}{*}{LOC-L}
      & \ding{51} & \ding{51} & \ding{55} & \ding{55} &8.99  &71.13  & 88.24 \\
      & \ding{51} & \ding{51} & \ding{55} & \ding{51} &14.21  &73.56  & 80.32 \\
      & \ding{51} & \ding{51} & \ding{51} & \ding{51} &18.79  &75.35  & 70.28 \\
    \bottomrule
  \end{tabular}
\end{table}
\begin{table}[t]
  \centering
  \caption{Replacing \(\mathcal{L}_{\text{DCL}}\) with other Loss Functions.. CosSim refers to cosine similarity.}
  \label{tab:loss}
  \begin{tabular}{ c| c c }
    \toprule
       Model & Loss Function & mIoU \\
    \midrule
    \multirow{3}{*}{LOC-F} 
      & CosSim         & 13.37 \\
      & CosSim w/ CB   & 14.41 \\
      & $\mathcal{L}_{\text{DCL}}$ & 35.10 \\
    \midrule
    \multirow{2}{*}{LOC-L} 
      & CosSim &  12.18\\
      & $\mathcal{L}_{\text{DCL}}$ &18.79  \\
    \bottomrule
  \end{tabular}
\end{table}

\begin{figure*}[ht]
        \centering
        \includegraphics[width=0.9\linewidth]{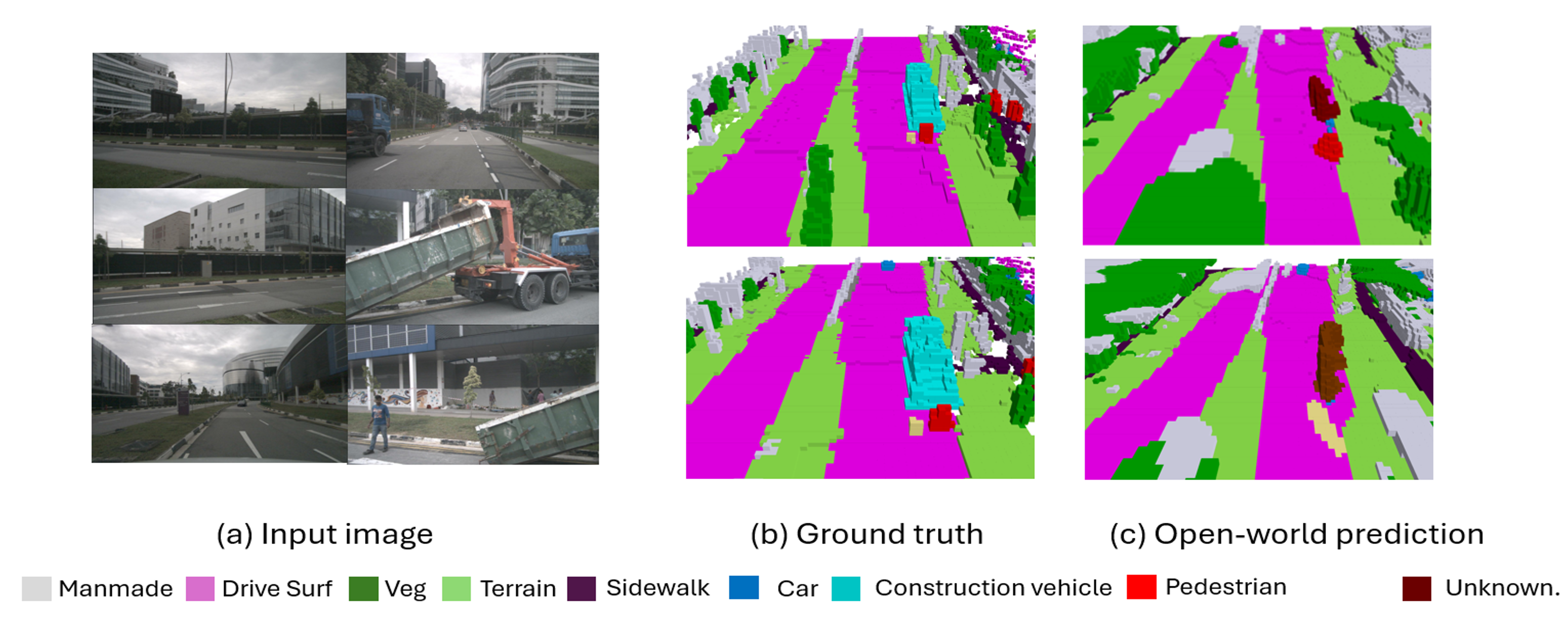}
        \vspace{-3mm}
        \caption{Results from the nuScenes dataset, where the unknown class is the construction vehicle}
        \label{fig：result}
\end{figure*}
\subsection{Comparison }
 This section comprehensively demonstrates the generality of the LOC framework through experiments with multiple networks, validating its effectiveness in both self-supervised and supervised tasks. We first analyze its open-set capabilities, then provide a rigorous performance comparison with existing state-of-the-art models on known classes.\par
\paragraph{Open-Set Settings.} We refer to methods \cite{hendrycks2016baseline,ming2022delving, logitnorm} from the open-set 2D semantic segmentation domain and implement these methods on the 3D occupancy prediction task, treating them as baselines for comparison. Closed-set methods do not consider unknown classes and thus do not evaluate open-set metrics. We set up multiple groups of unknown classes to comprehensively demonstrate the efficacy of our approach. When calculating the mIoU metric, these unknown classes are ignored. Experimental results in Table \ref{tab:main} show that, compared to the baseline model, the mIoU metric indicates that our approach does not compromise the classification ability for known classes while being able to distinguish unknown classes, achieving better results on open-set evaluation metrics. We show qualitative results of our approach in Fig. \ref{fig：result}.\par
\paragraph{Comparison with occupancy predictions.}
In Table \ref{tab:state}, the first six rows of  Table 1 list non-language-driven prediction models, including TPVFormer \cite{huang2023tri}, OccFormer \cite{zhang2023occformer}, BEVFormer \cite{li2022bevformer}, and BEVDet \cite{huang2021bevdet}, as well as the self-supervised SelfOcc \cite{huang2024selfocc}. The remaining rows present language-driven occupancy models.. As shown in the table, our model achieves an mIoU of 35.10. Our model demonstrates competitive performance compared to other supervised occupancy models.
We also compared our approach with state-of-the-art open-vocabulary methods, such as VEON \cite{zheng2025veon} and LangOcc \cite{boeder2024langocc}.
The results indicate that our model outperforms other language-driven models, which can be attributed to the introduction of DCL that more effectively utilizes limited annotated data, outputting denser text-aligned features. Simultaneously, our models based on BEVDet and FlashOcc demonstrate competitive performance compared to other occupancy estimation approaches. We note that although the performance of models based on TPVFormer might be lower than others, our LOC framework is capable of achieving performance close to that of occupancy networks under supervised conditions.

\subsection{Ablation Study}

\paragraph{Different training modules.} In Table \ref{tab:ablation}, we provide ablation studies to investigate the contribution of the modules we introduced. Among them, LOC-F already uses GT, so the densification strategy is not needed. As shown in the 5th row of Table 3, when the DCL component is not employed in LOC-L, it implies directly assigning image features to the nearest occupied voxels via nearest neighbor for direct distillation. This direct distillation approach often leads to feature over-homogenization and a significant loss of feature diversity, which in turn results in substantial open-set performance issues. Furthermore, it simultaneously incurs considerable computational overhead for distilling dense high-dimensional features. Therefore, \(\mathcal{L}_{\text{DCL}}\) is one of the critical modules to make our work effective.
\paragraph{Replacing \(\mathcal{L}_{\text{DCL}}\) with other Loss Functions.} In Table \ref{tab:loss}, we evaluate the impact of replacing our proposed \(\mathcal{L}_{\text{DCL}}\) with other alternatives: Cosine Similarity loss, Cosine Similarity loss with Class Balance. When \(\mathcal{L}_{\text{DCL}}\) is replaced with standard Cosine Similarity loss, model performance decreased, attributed to class imbalance in the nuScenes dataset, which biases the model toward learning features of majority classes (those with larger sample sizes). Notably, even when using Cosine Similarity loss with Class Balance, the performance gain is marginal and fails to mitigate the issue effectively.  These results validate that our proposed \(\mathcal{L}_{\text{DCL}}\) is critical for generating dense, text-aligned voxel features.

\section{Conclusion}
 This paper proposes a novel and general LOC framework, aiming to address the challenge of open-set 3D occupancy prediction in autonomous driving. The core contribution lies in our designed Robust Densification Strategy, which effectively solves the problems of sparse 3D data and voids, generating high-quality dense occupancy representations. Building upon this, we introduce Dense Contrastive Learning, which effectively elevates 2D vision-language information into 3D space by aligning dense voxel features with CLIP text embeddings. DCL avoids feature over-homogenization and performance overhead caused by directly distilling high-dimensional features, while also possessing the potential to further enhance performance by leveraging existing 3D occupancy ground truth. Our framework provides a new paradigm, enabling further extensions in open-set 3D occupancy prediction.

\noindent {\bf Imapact and Limitation}. Beyond autonomous driving, the proposed 3D occupancy prediction model has diverse applications. In VR/AR, it enriches virtual experiences and aids in industrial AR inspections. For robotics and embodied AI, it improves navigation for delivery robots. In smart homes, it enables intelligent environmental control by analyzing room occupancy, enhancing energy efficiency. Despite the notable achievements of our proposed method, several limitations exist. First, in complex scenarios with numerous objects and occlusions, like crowded urban intersections, the model can be improved by spatial-relationship understanding. Second, in dynamic scenes where video objects are in continuous motion, such as in busy traffic, the model can be improved by long-term modeling. Third, its performance is closely tied to the quality of pre-trained VLM, whose biases or limited generalization can lead to inaccurate predictions.

\section*{Acknowledgements}

Funding for this work was provided in part by the HUST Interdisciplinary Research Support Program (2025JCYJ077), and the 2026 Optics-Valley Excellence Project funded by the National Graduate College for Elite Engineers of HUST.

\bibliography{aaai2026}

\newpage

\section{Appendix}

\subsection{Vocabulary}
We present the vocabulary utilized for semantic occupancy estimation on the Occ3D-nuScenes\cite{caesar2020nuscenes} dataset and for the training of our model. For Group A, we adhere to the original nuScenes categories, while for Group B, we map the original 16 classes to an expanded set of 43 classes, as detailed in the accompanying table \ref{tab:mapping}. For each category within the Occ3D-nuScenes benchmark\cite{tian2024occ3d}, we have established a collection of textual prompts that delineate the respective class. We employ a straightforward prompt engineering technique prior to extracting CLIP text features from a set of text prompts. For each object class "XX" (excluding the "other" and‘ free’ class), we reformulate the text prompts as "a XX in a scene," such as "a car in a scene." we juxtapose the estimated voxel features against each textual embedding from our vocabulary, assigning to each voxel the label associated with the prompt that garners the highest similarity score.

\begin{table}[h]
    \centering
    \begin{tabular}{c c| c c c}
        \toprule
        {$\tau_1$} & {$\tau_2$} & {mIoU} & AUPR\(\uparrow\) & FPR95\(\downarrow\) \\
        \midrule
        0.01  & 0.5  & 34.8  & 78.18 & 65.63 \\
        0.08  & 0.5  & 35.01 & 79.04 & 66.27 \\
        0.2   & 0.5   & 34.93 & 79.63 & 64.79 \\
        0.5   & 0.08   & 35.10  & 80.64 & 64.88 \\
        0.5   & 0.2   & 35.10  & 80.64 & 64.88 \\
        0.5   & 0.5   & 35.10  & 80.42 & 63.83 \\
        0.5   & 2   & 35.10  & 80.42 & 63.85 \\
        4     & 0.5     & 34.47 & 81.18 & 62.60 \\
        4     & 4     & 34.47 & 81.18 & 62.59 \\
        \bottomrule
    \end{tabular}
    \caption{Performance metrics under different $\tau_1$ and $\tau_2$}
    \label{tab:tau_performance}
\end{table}

\begin{table*}[ht]
    \centering
    \begin{tabular}{c |c}
        \hline
        \textbf{Original Categories (Group A)} & \textbf{Mapped Category (Group B)} \\
        \hline
        barrier               & barrier, barricade \\
        bicycle               & bicycle \\
        bus                   & bus \\
        car                   & car \\
        construction vehicle  & bulldozer, excavator, concrete mixer, crane, dump truck \\
        motorcycle            & motorcycle \\
        pedestrian            & pedestrian, person \\
        traffic cone          & traffic cone \\
        trailer               & trailer, semi trailer, cargo container, shipping container, freight container \\
        truck                 & truck \\
        drivable surface      & road \\
        other flat            & curb, traffic island, traffic median \\
        sidewalk              & sidewalk \\
        terrain               & grass, grassland, lawn, meadow, turf, sod \\
        manmade               & building, wall, pole, awning \\
        vegetation            & tree, trunk, tree trunk, bush, shrub, plant, flower, woods \\
        \hline
    \end{tabular}
    \caption{Mapping from Group A to Group B. Here we list the 43 pre-defined class names corresponding to the 16 nuScenes classes}
    \label{tab:mapping}
\end{table*}

\begin{figure}[ht]
    \centering
    \includegraphics[width=0.5\textwidth]{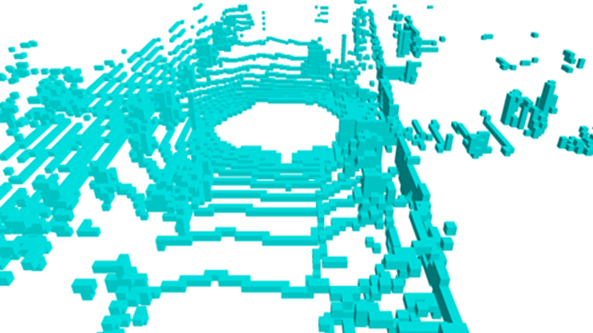} 
    \caption{3D feature space sparsity }
    \label{fig:dense}  
\end{figure}

\begin{table*}[ht]
    \centering
    \footnotesize  
    \begin{tabular}{p{1.0cm}| p{0.4cm} p{0.4cm} p{0.4cm} p{0.4cm} p{0.4cm} p{0.4cm} p{0.4cm} p{0.4cm} p{0.4cm} p{0.4cm} p{0.4cm} p{0.4cm} p{0.4cm} p{0.4cm} p{0.4cm} p{0.5cm}| p{0.5cm} p{0.5cm}}
    \toprule
         Prompt & 
        \rotatebox{90}{\raisebox{-0.2em}{\textcolor{orange}{\rule{0.5em}{0.5em}}}\, barrier} & 
        \rotatebox{90}{\raisebox{-0.2em}{\textcolor{pink}{\rule{0.5em}{0.5em}}}\, bicycle} & 
        \rotatebox{90}{\raisebox{-0.2em}{\textcolor{yellow}{\rule{0.5em}{0.5em}}}\, bus} & 
        \rotatebox{90}{\raisebox{-0.2em}{\textcolor{blue}{\rule{0.5em}{0.5em}}}\, car} & 
        \rotatebox{90}{\raisebox{-0.2em}{\textcolor{cyan}{\rule{0.5em}{0.5em}}}\, cons. veh.} & 
        \rotatebox{90}{\raisebox{-0.2em}{\textcolor{orange}{\rule{0.5em}{0.5em}}}\, motorcycle} & 
        \rotatebox{90}{\raisebox{-0.2em}{\textcolor{red}{\rule{0.5em}{0.5em}}}\, pedestrian} & 
        \rotatebox{90}{\raisebox{-0.2em}{\textcolor{yellow}{\rule{0.5em}{0.5em}}}\, traffic cone} & 
        \rotatebox{90}{\raisebox{-0.2em}{\textcolor{brown}{\rule{0.5em}{0.5em}}}\, trailer} & 
        \rotatebox{90}{\raisebox{-0.2em}{\textcolor{purple}{\rule{0.5em}{0.5em}}}\, truck} & 
        \rotatebox{90}{\raisebox{-0.2em}{\textcolor{pink}{\rule{0.5em}{0.5em}}}\, drive. surf.} & 
        \rotatebox{90}{\raisebox{-0.2em}{\textcolor{red}{\rule{0.5em}{0.5em}}}\, other flat} & 
        \rotatebox{90}{\raisebox{-0.2em}{\textcolor{purple}{\rule{0.5em}{0.5em}}}\, sidewalk} & 
        \rotatebox{90}{\raisebox{-0.2em}{\textcolor{green}{\rule{0.5em}{0.5em}}}\, terrain} & 
        \rotatebox{90}{\raisebox{-0.2em}{\textcolor{white}{\rule{0.5em}{0.5em}}}\, manmade} & 
        \rotatebox{90}{\raisebox{-0.2em}{\textcolor{green}{\rule{0.5em}{0.5em}}}\, vegetation} & 
        \multicolumn{1}{c}{mIoU} \\
    \midrule
        A  & 39.02 & 13.16 & 35.18 & 43.97 & -- & 10.07 & 8.46 & 18.94 & 26.41 & 30.64 & 78.22 & 36.17 & 48.05 & 50.5 & 34.85 & 29.91 & 33.57 \\
        B & 39.2 & 14.31 & 34.99 & 44.03 & -- & 17.3 & 16.53 & 18.07 & 27.32 & 30.41 & 79.06 & 38.03 & 48.47 & 51.84 & 35.16 & 30.38 & 35.01 \\
        C & 39.28 & 12.84 & 36.96 & 44.9 & -- & 15.32 & 17.42 & 18.89 & 27.87 & 31.55 & 78.92 & 36.66 & 47.97 & 51.21 & 35.89 & 30.85 & 35.10 \\
    \bottomrule
    \end{tabular}
    \caption{Performance of different prompting strategies. A, B, and C represent different prompting strategies. ‘-’ indicates unknown classes that are not predicted during inference.}
    \label{tab:iou}
\end{table*}

\subsection{Feature space sparsity}
In Figure \ref{fig:dense}, we demonstrate the results of projecting 2D pixel coordinates into 3D space solely based on LiDAR information (due to the relative accuracy of LiDAR data). It can be observed that the 3D feature space is highly sparse, with most voxel spaces lacking corresponding 2D information, which poses significant challenges for feature distillation. Additionally, we considered the forward projection approach. While this method can alleviate sparsity to some extent, the results remain quite discrete and heavily rely on the accuracy of the depth estimation model. In occupancy prediction models, BEV  encoders are typically used to densify features through convolutional operations, whereas backward projection may introduce substantial mismatching issues.

\begin{table}[h]
    \centering
    \renewcommand{\arraystretch}{1.3} 
    \begin{tabular}{l|cc}
        \hline
        \textbf{Method}  & \textbf{AUPR$\uparrow$} & \textbf{FPR95$\downarrow$} \\
        \hline
        MCM 0.08       & 0.7198         & 0.8323         \\
        MCM1           & 0.7175         & 0.8126         \\
        MCM 0.01       & 0.6257         & 0.8670         \\
        MCM 5          & 0.7180         & 0.8109         \\
        MCM 15         & 0.7170         & 0.8083         \\
        MSP            & 0.7248         & 0.8131         \\
        \hline
    \end{tabular}
    \renewcommand{\arraystretch}{1} 
    \caption{Comparison of additional baseline results. The temperature parameter is denoted by the value following each method name (e.g., MCM 0.08 uses a temperature of 0.08).}
    \label{fig:appendix-baseline}
\end{table}

\subsection{Additional Experiments}
\paragraph{Different hyper-parameter.} We tested the impact of different contrastive loss temperature values \( \tau_1 \) and \( \tau_1 \) on the model’s performance in Table \ref{tab:tau_performance}. We observe that increasing the \( \tau \) value leads to an improvement in Mean AUROC and a reduction in FPR95, indicating better discrimination between known and unknown categories. However, the mIoU metric shows minimal variation across different \( \tau \) values, an excessively high temperature value may slightly impair the model's ability to predict known classes accurately. Conversely, the impact of \( \tau_2 \) on the model's performance is not significant.

\paragraph{Additional Baseline Results}
In this section, we present more baseline results in Table \ref{fig:appendix-baseline}. These results help to better understand the performance of different approaches under our open-world settings.We evaluate all methods on the nuScenes dataset, using the standard training and validation splits. The evaluation metrics include Mean AUROC and Mean FPR95, which are commonly used for out-of-distribution detection tasks. 

\noindent\textbf{Feature Alignment with CLIP.}
 We explore the alignment of the model outputs with CLIP text embeddings. We test the model's performance on the closed set using different prompts, with the results shown in Table \ref{tab:iou}. Group A uses 16 classes from the nuScenes dataset as prompts. Due to the relatively vague labels in the dataset, we performed simple replacements, such as replacing "manmade" with "building" and "drivable surface" with "road." Detailed changes are provided in the appendix. Group B combines the original 16 classes and maps them to 43 classes, while Group C employs a simple prompt engineering trick, "a label in a scene," which aligns with most of our experiments. The final results show that Group A slightly reduces the mIoU performance, while Groups B and C yield almost same results. This validates that our model successfully outputs dense voxel features aligned with the CLIP text feature space, as these prompts should have similar features in the CLIP feature space.

\end{document}